\documentclass[sigconf]{acmart}
\usepackage{enumitem}

\AtBeginDocument{%
  \providecommand\BibTeX{{%
    \normalfont B\kern-0.5em{\scshape i\kern-0.25em b}\kern-0.8em\TeX}}}

\copyrightyear{2020} 
\acmYear{2020} 
\setcopyright{rightsretained} 
\acmConference[CIKM '20]{Proceedings of the 29th ACM International Conference on Information and Knowledge Management}{October 19--23, 2020}{Virtual Event, Ireland}
\acmBooktitle{Proceedings of the 29th ACM International Conference on Information and Knowledge Management (CIKM '20), October 19--23, 2020, Virtual Event, Ireland}\acmDOI{10.1145/3340531.3412137}
\acmISBN{978-1-4503-6859-9/20/10}

\begin{document}

\title{Ranking Clarification Questions via Natural Language Inference}


\author{Vaibhav Kumar}
\authornote{Both authors contributed equally to this research.}
\affiliation{\institution{Carnegie Mellon University}}
\email{vaibhav2@andrew.cmu.edu}

\author{Vikas Raunak}
\authornotemark[1]
\affiliation{\institution{Carnegie Mellon University}}
\email{vraunak@andrew.cmu.edu}

\author{Jamie Callan}
\affiliation{\institution{Carnegie Mellon University}}
\email{callan@cs.cmu.edu}

\renewcommand{\shortauthors}{Kumar and Raunak, et al.}

\begin{abstract}
Given a natural language query, teaching machines to ask clarifying questions is of immense utility in practical natural language processing systems. Such interactions could help in filling information gaps for better machine comprehension of the query. For the task of ranking clarification questions, we hypothesize that determining whether a clarification question pertains to a missing entry in a given post (on QA forums such as StackExchange) could be considered as a special case of Natural Language Inference (NLI), where both the post and the most relevant clarification question point to a shared latent piece of information or context. We validate this hypothesis by incorporating representations from a Siamese BERT model fine-tuned on NLI and Multi-NLI datasets into our models and demonstrate that our best performing model obtains a relative performance improvement of 40 percent and 60 percent respectively (on the key metric of Precision@1), over the state-of-the-art baseline(s) on the two evaluation sets of the StackExchange dataset, thereby, significantly surpassing the state-of-the-art. 
\end{abstract}



\maketitle

\section{Introduction}

The advent of conversational systems like “Alexa”, “Siri” etc. has led to the identification of various interesting problems at the intersection of Natural Language Processing (NLP) systems and human interaction. 
One such problem is to ask clarifying questions when a system is presented with a natural language query. 
According to \citet{rao2018learning}, the primary goal of a clarification question is to fill in the information gaps. 
This can be explained via the following example: suppose there is an underspecified (or ambiguous) post $p$ on a QA forum. 
Then, a good clarification question $q$, for $p$, would be the one whose answer leads to a resolution of the underspecification associated with $p$. 
In more simpler terms, a good question will be the one whose likely answer will be useful.

\citet{rao2018learning} pose the problem of coming up with clarification questions in an Information Retrieval setting, where the task is to retrieve clarifying questions from a pool of candidate questions with respect to a particular post. 
They propose a model based on the decision theoretic principle of \textit{Expected Value of Perfect Information} (EVPI). 
Further, in order to facilitate training, \citet{rao2018learning} have provided a dataset based on three domains of StackExchange, which provides triplets of $(p, q, a)$, where $p$ is the original post, $q$ is the clarification question and $a$ is the most probable answer to the clarification question.

Based on the definition of a clarification question (as provided earlier), we can ascertain that the principal modeling objective of the problem is that a clarification question $q$ should help in increasing the probability of the correct answer to the given post $p$. 
An underlying assumption behind this statement is that the context of the answer will be useful in determining the most relevant question. 
In this work, we study this modeling problem as well as test its assumptions from a representation learning point of view. 
More specifically, we hypothesize and validate that ranking clarification questions could be treated as a special case of a Natural Language Inference task.




\section{Related Work}

There are only a couple of datasets that explicitly contain clarification questions: \citet{rao2018learning} and ClarQ \cite{kumar-black-2020-clarq}. We review two lines of related work: (1) Ranking of Clarification Questions and (2) Transfer Learning.

\textbf{Ranking of Clarification Questions}:  There hasn't been a lot of work on this problem. 
Recently, \citet{aliannejadi2019asking} formulated the task of asking clarifying questions in open-domain information-seeking conversational systems, and provided a dataset, Qulac. 
They posit that a system can improve user satisfaction by proactively asking questions that can help satisfy users' information needs. 
Additionally, they proposed a retrieval framework consisting of three components: question retrieval, question selection, and document retrieval. 
However, the dataset consists of keyword based queries and does not resemble the problems posed by natural language questions.

The work which is most relevant to ours is that of \citet{rao2018learning}, which proposes a neural ranking model inspired by the theory of Expected Value of Perfect Information (EVPI) \citep{avriel1970value} to rank clarification questions. 
The EVPI model is decomposed into two components (which is modelled jointly): (1) Answer Modelling and (2) Utility Calculator. 
The answer modelling component tries to assess the value of an answer associated with a post and a candidate clarification question. 
On the other hand, the utility calculator computes the utility of concatenating the post with the answer to the candidate clarification question.

\textbf{Transfer Learning}: The use of large-scale pre-trained models such as BERT \cite{devlin2018bert} for transfer learning has become prevalent in several Information Retrieval tasks. 
A number of recent works have further tried to improve upon the generic transfer learning methods of using representations and fine-tuning, and have proposed to further adapt BERT to their tasks, using additional training tricks such as multi-task learning and contrastive learning. 
In particular, \citet{rosset2020leading} utilize a multi-task fine-tuning approach for the task of query suggestion (in context of ad-hoc search) and demonstrate that multi-tasking helps in improving the overall $CTR$ of search results.

\section{Proposed Methods}

We formally describe the problem as follows: Given an initial post $p_i \in P$ and a set of candidate clarification questions $q_i \in Q$, the task is to provide a ranked list of $N$ clarification questions $[q_{i1}, q_{i2} \dots q_{iN}] \forall q_{ij} \in Q$, such that the more relevant questions are higher up in the ranked list. 
Each $q_ij$ is also paired with an answer $a_ij$ which can be used as an additional signal for model training. 
Note that $a_ij$ is an answer to $q_ij$ and not the post $p_i$ \footnote{In all our experiments, $N$ is set to 10.}. 


Before describing our proposed methods, we draw an analogy between the task of natural language inference and the pair regression task of ranking clarification questions.

\textbf{Natural Language Inference (NLI) }: The task of natural language inference is to determine whether, given a premise, a hypothesis can be inferred. 
The task has been widely studied in the Natural Language Processing community, where two large scale datasets, namely the Stanford Natural Language Inference (SNLI) corpus \cite{snli} and the Multi-Genre NLI (MultiNLI) corpus \cite{multinli} have been made available to evaluate different representations and models \cite{wang2018glue}.

\textbf{Task Analogy with NLI }: \label{rep3} In the task of evaluating a candidate clarification question for a given post, the classifier has to predict whether the question follows from the context of the post. This formulation can be juxtaposed to the task of natural language inference (NLI), where, given a premise, the classifier has to predict whether a hypothesis is true (entailment), false (contradiction) or neutral. However, the task of evaluating a candidate clarification question is narrower in scope. More precisely, the task of of evaluating a candidate clarification question for a given post is a binary inference problem, where the inference is made on the basis of latent information that is common to the context of both the question and the post. Consequently, we hypothesize that our problem is a special case of the natural language inference (NLI) task, since, given a post and a question, the problem is to determine whether the two share any common latent information, based on the context. 

\textbf{Leveraging NLI }: The previous analogy suggests that pre-training the classifier on NLI datasets might be useful to bootstrap the model for ranking clarification questions. To this end, we use the Sentence-BERT models (SBERT) \citep{sbert} pretrained on SNLI \citep{snli} and MultiNLI \citep{multinli} datasets. Using the SBERT model to extract representations for the posts, questions and answers, we build the following models:
\begin{enumerate}[leftmargin=*]
    \item \textbf{SBERT-PQ}: We extract post (P) and question (Q) representations (768 dimensional) from SBERT and feed it to a three layer fully connected network, before outputting softmax probabilities, of the question being true (relevant) or false (non-relevant). The loss function is the standard cross entropy loss. 
    \item \textbf{SBERT-PQA}: Besides the post (P) and question (Q), we also extract representations for the corresponding answer (A) from SBERT (768-dimensional). Rest of the model is the same as SBERT-PQ.
    \item \textbf{SBERT-Large-PQ}: We use SBERT-Large to extract the representations (1024-dimensional) for the posts (P) and questions (Q). Rest of the model is the same as SBERT-PQ.
    \item \textbf{SBERT-Large-PQA}: Besides the post (P) and question (Q), we also extract representations for the corresponding answer (A) from SBERT-Large (1024-dimensional). Rest of the model is the same as SBERT-Large-PQ.
\end{enumerate}

In all the SBERT models, we extract the mean pooled final-layer representations of the tokens.

\section{Experimental Details}

\subsection{Dataset}
The data in \citet{rao2018learning} was collected using posts on three different domains of StackExchange, namely askubuntu, unix and superuser. The training set consists of 61678 triples of $(p, q, a)$ and the validation set consists of $7710$ triples of $(p, q, a)$. Each $(p_i, q_i, a_i)$ triple is paired with the top 9 $(p_j, q_j, a_k)$ triples retrieved using Lucene. Note that the retrieval is performed by matching the content of the posts and not their questions and answers. Thus, each post consists of 10 possible candidate questions (and the answers to these questions) which need to be ranked.

The test set consists of 500 posts whose candidate questions have been human annotated. The annotations are of two types:
\begin{enumerate}[leftmargin=*]
    \item\textbf{Best annotation}: Each annotator is asked to mark the single best question that can be asked. Finally, per post, the union of the best annotations of different annotators are used as the ground truth.
    \item \textbf{Valid annotation}: Each annotator is asked to mark all valid questions that could have been asked (more than 1). Finally, per post, the intersection of the valid annotations of different annotators are used as ground truth.
\end{enumerate}
Our models are evaluated on both these types of annotations. We utilize precision as our evaluation metric. More specifically, we measure precision at top 5 positions in the ranked list. One must also note that, we are primarily more concerned with $P@1$. This is because the main downstream application of ranking clarification question will be for the task of conversational question-answering where the system will be limited to provide only a single response.

\subsection{Baselines}
Apart from the EVPI and PQA model described in \cite{rao2018learning}, we experiment with two other types of models.

\subsubsection{Bidirectional PQA Model}
\label{rep1}
The model utilises three separate Bi-LSTMs for generating a representation for the post, question and answer respectively. Thus, given a $(p, q , a)$ triple, the model first generates a representation $(\Bar{p}, \Bar{q}, \Bar{a})$ using the Bi-LSTM. It later concatenates the representations and utilizes it as an input to a fully-connected feed forward network. The feed-forward network consists of 10 layers with ReLU activations. The model is trained using the cross-entropy loss.


\subsubsection{Using Contextualized Representations}
\label{rep2}
A natural extension to the Bidrectional PQA Model (as proposed above) is to utilise BERT \citep{devlin2018bert} as it has shown to be capable of effectively modelling longer sequences and producing more meaningful representations.

For our experiments, we fine-tune DistilBERT \citep{sanh2019distilbert} which is a distlilled version of the pre-trained BERT-base model. We use it in the following two settings:

\noindent \textbf{DistilBERT-PQA} : In this setting we concatenate the post (p), clarification question (q) and answer (a) and use it as an input to BERT. We use the representation of the [CLS] token obtained in the final layer and further pass it to a fully connected layer. The model is then fine-tuned using the cross-entropy loss.

\noindent  \textbf{DistilBERT-PQ} : In contrast to DistilBERT-PQA, here we do not use the answers. All other settings remain the same.

The problem of ranking questions in this case is posed as a binary classification problem i.e. a question belongs to class 1 if it a good clarification question with respect to a given post, else its classified as 0. At test time, in order to create a ranked list, we sort the questions based on a decreasing order of their probability of belonging to class 1.


\subsection{Parameter Settings}
The embedding layer of bidirectional PQA Model is initialized using 200 dimensional Glove embeddings pre-trained on the StackExchange dataset. The hidden dimension of the LSTM and the feedforward networks are set to 200. The model is trained with a batch size of 120 and the learning rate is set to 0.001. We fine-tune DistilBERT based models for 5 epochs. The batch size is set to 16 and a learning rate warmup of 1920 steps is used. We use ADAM \citep{kingma2014adam} as the optimizer with a learning rate of 3e-6. For all SBERT based models, the batch size was set to 1000, and the models were trained for 50 epochs with a learning rate of 0.01 with the Adam optimizer. Dropout (0.4) was applied before each of the three linear layers. Since, in the case of SBERT, we first extracted the representations and then consumed them in our models, we effectively use frozen SBERT models (no-fine-tuning).

\section{Results and Discussion}
The key metric to compare the performance of the various models is \textbf{P@k}, which refers to Precision of the top-k results generated by the model. Tables \ref{tab:best} and \ref{tab:valid} describe the performance of the various models, along with the results reported from \citet{rao2018learning}. \citet{rao2018learning} only report Precision at 1, 3 and 5, while we also report the Precisions at 2 and 4 in both the tables.

We make the following observations based on the obtained results:
\begin{enumerate}[leftmargin=*]
    \item The \textbf{Bidi-PQA} model significantly outperforms the EVPI model, especially on P@1 and P@5 metrics. 
    \item The performance of all \textbf{DistilBERT} based models is lower than that of PQA, EVPI as well as Bidi-PQA. This clearly suggests that using the representations of DistilBERT was detrimental for the task at hand. One of the possible reasons could be that StackExchange posts consist of code snippets and character patterns that need to be recognized in an effective manner. Perhaps, off-the-shelf BERT (when it receives out-of-domain data) was ineffective in identifying such patterns as most of it gets treated as unseen tokens. On the other hand, since Bidi-PQA and EVPI utilize Glove embeddings pre-trained on the StackOverflow data, they are able to perform much better. Our finding here is consistent with a very recent work by \citet{tabassum2020code}. \citet{tabassum2020code} show that Glove embeddings perform far better than a fine-tuned off-the-shelf BERT on the task of named entity recognition on the StackOverflow data. They also show that training BERT on StackOverflow data (with the use of a special tokenizer) achieves much better performance than a fine-tuned off-the-shelf BERT. We believe that adopting a similar strategy would be beneficial for our task as well, given the domain mismatch.
    
    \item All SBERT based models are in par or better than EVPI in terms of P@1 scores. In fact, \textbf{SBERT-PQ} model (which is the best performing SBERT model) provides a 60\% improvement over EVPI when tested using Best Annotations and a 40\% improvement over EVPI when tested using Valid Annotations (in terms of P@1 scores). The P@1 performance of SBERT-PQ model is also better than that of Bidi-PQA. The task that we are entailed with is a high precision task, and thus a model which has a higher P@1 can be considered to be a better model overall. The better performance of SBERT model also validates our \textbf{hypothesis} (Section \ref{rep3}) that the problem of ranking clarification questions is a special case of a natural language inference task.  In order to predict a good clarification question, the model should be able to infer the latent information that is common to the context of both the question and the post. We hypothesize that this is the core source of the gains observed in this case.
    
    \item When comparing the PQ and PQA counterparts of the DistilBERT and SBERT models, we observe that \textbf{PQ models perform better than PQA models}. This means that a majority of the times, incorporating answers within the model was detrimental to the performance. We hypothesize that this could be an artifact of the data itself. \citet{rao2018learning} identify answers to each questions based on certain heuristics like time stamps, edit history etc. It could be very well possible that the answers might not be actual answers at all. For certain models, like SBERT, such answers might be causing a drift in the models understanding of a good clarification question i.e owing to lower semantic similarity, answers could be acting as noise.
    
\end{enumerate}

Finally, regarding the overall performance of the various methods, we can observe that the SBERT based models generally obtain the highest P@1 scores thereby producing a new state-of-the-art for the given task.

\begin{table}[ht!]
    \centering
\begin{tabular}{l|l|l|l|l|l}
\textbf{Model}      & \textbf{P@1}  & P@2  & \textbf{P@3}  & P@4  & \textbf{P@5}  \\ \hline
Bidi-PQA   & \textbf{28.6} & 25.7 & 25.3 & 24.5 & 23.9 \\ \hline
DistilBERT-PQ & 18.2 & 20.1 &20.1 & 19.9& 20.5 \\
DistilBERT-PQA & 24.4 & 21.4 & 20.7 & 21.0 & 21.0\\ \hline
SBERT-PQ  & \textbf{44.4} & 28.1   & 22.93 & 21.05   & 19.56 \\
SBERT-PQA  & 28.4 & 24.7   & 22.79 & 22.05   & 21.52 \\
SBERT-Large-PQ  & 42.6 & 27.1   & 21.86 & 19.05   & 17.96 \\
SBERT-Large-PQA & 40.4 & 26.8   & 21.66 & 19.6   & 18.16 \\ \hline
PQA  & 25.2 & NA   & 22.7 & NA   & 21.3 \\
EVPI & 27.7 & NA   & 23.4 & NA   & 21.5 \\ \hline
\end{tabular}
\caption{Results of the methods on the best annotations of ground truth. PQA and EVPI results are from \cite{rao2018learning}.}
\label{tab:best}
\end{table}
\vspace{-0.7em}
\begin{table}[ht!]
    \centering
   \begin{tabular}{l|l|l|l|l|l}
\textbf{Model }     & \textbf{P@1}  & P@2  & \textbf{P@3}  & P@4  & \textbf{P@5}  \\ \hline
Bidi-PQA   & \textbf{37.8} & 32.9 & 32.5 & 31.8 & 31.5 \\ \hline
DistilBERT-PQ & 24.4 & 25.6 & 26.9 & 27.6 & 28.4 \\
DistilBERT-PQA & 31.6 & 28.0 & 28.0 & 28.4 & 29.0 \\ \hline
SBERT-PQ  & \textbf{50.6} & 36.2  & 31.8 & 29.15   & 27.44 \\
SBERT-PQA  & 36.6 & 33.2   & 31.93 & 30.8   & 30.2 \\
SBERT-Large-PQ  & 49.6 & 35.5   & 29.86 & 26.3   & 25.44 \\
SBERT-Large-PQA & 45.4 & 34.0   & 29.26 & 27.7   & 26.08 \\ \hline
PQA  & 34.4 & NA   & 31.8 & NA   & 30.1 \\ 
EVPI & 36.1 & NA   & 32.2 & NA   & 30.5 \\ \hline
\end{tabular}
\caption{Results of the methods on the Valid annotations of the ground truth. The results for PQA and EVPI have been copied from \cite{rao2018learning}.}
\label{tab:valid}
\end{table}

\section{Error Analysis}
\begin{figure}
    \centering
    \includegraphics[width=0.4\textwidth]{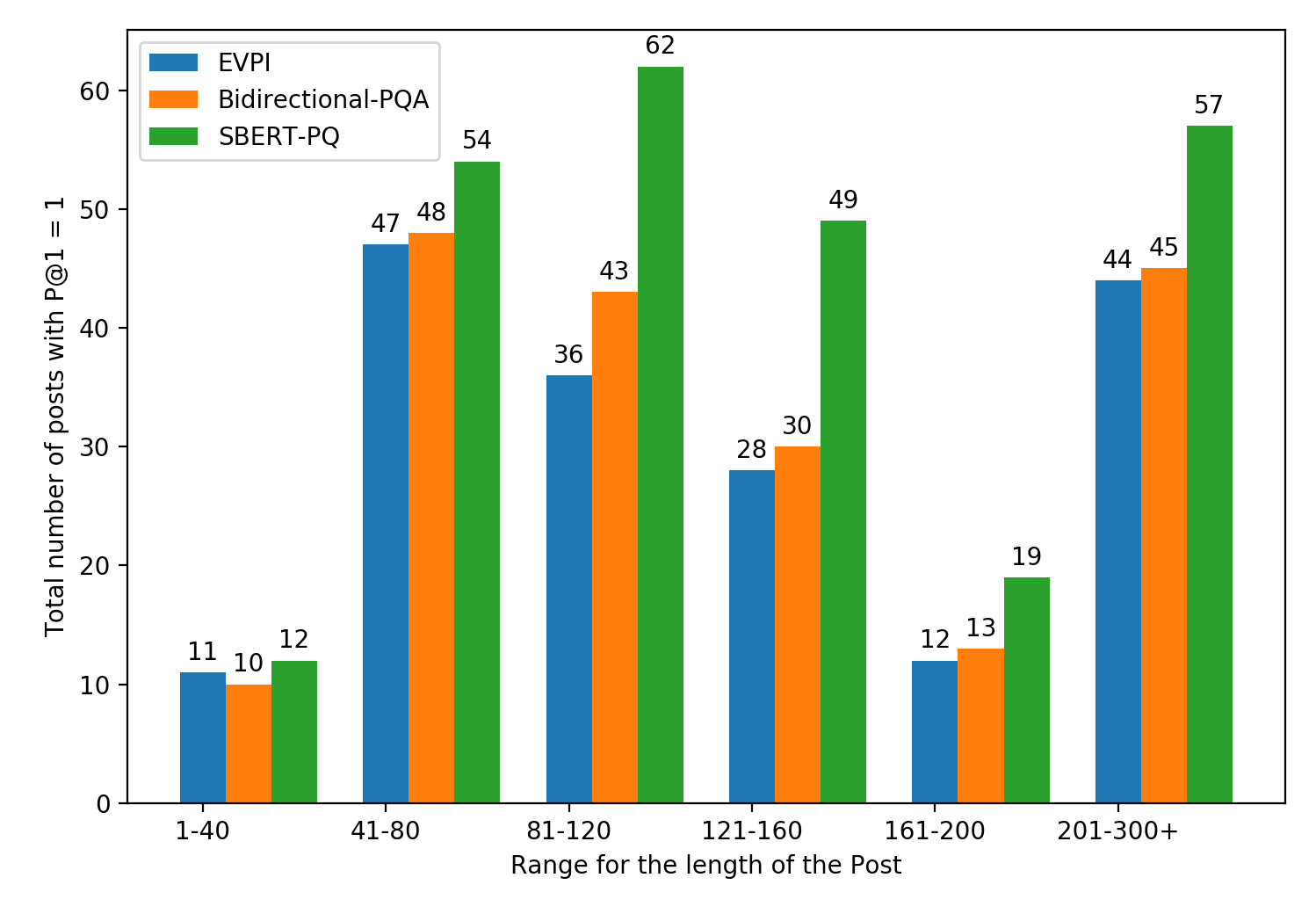}
    \caption{Number of Correctly Identified Clarification Questions (based on Valid Annotations) grouped by Post Length. In total, there are 23 posts with lengths between 1-40, 103 with length between 41-80, 114 with lengths between  81-120, 88 with length between 121-160 between, 44 with length between 161-200 and 128 with length between 201-300+.}
    \label{fig:figure1}
\end{figure}

\begin{figure}
    \centering
    \includegraphics[width=0.4\textwidth]{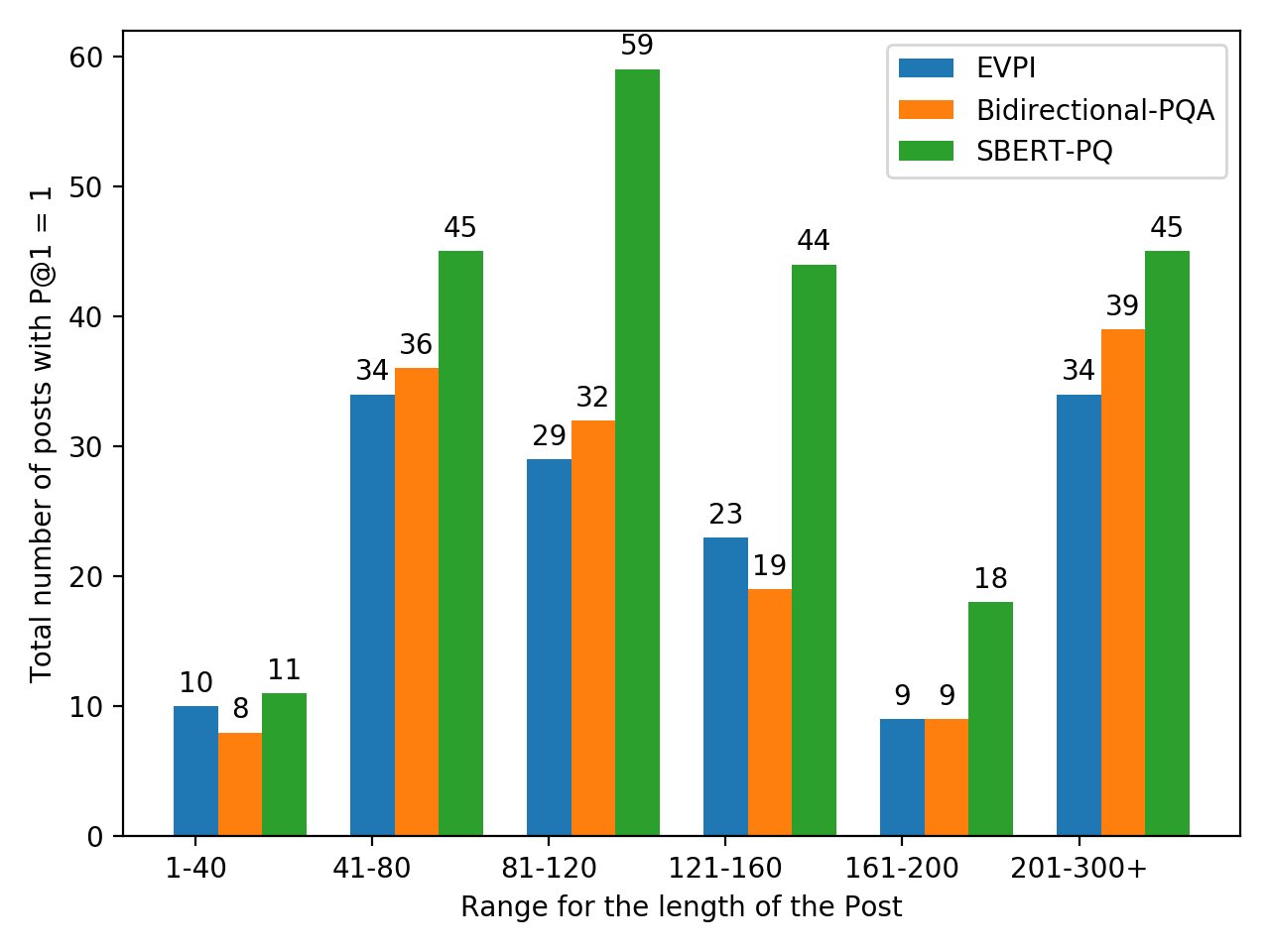}
    \caption{Number of Correctly Identified Clarification Questions (based on Best Annotations) grouped by Post Length.}
    \label{fig:figure2}
\end{figure}

Figure \ref{fig:figure1} shows the plot of the number of posts (bucketed by post length), for which each of the models got the correct clarification question at rank 1, when evaluated on the `Valid' evaluation set. The models compared are: EVPI (the published state-of-the-art), Bidirectional-PQA and SBERT-PQ (our new state-of-the-art). From the figure, it is evident that for posts in buckets of length 81 or larger, the SBERT-PQ model significantly outperforms the other two. This is despite the fact that it uses no answer information (unlike the other two). Although, it is hard to disentangle whether the gains arrive due to better representations or due to better inference of the shared latent information, given that, at shorter lengths (upto 40), EVPI does almost as well as SBERT-PQ. This also shows that without any NLI-based representations, the inference task becomes significantly harder for longer length posts, where SBERT-PQ gains a clear advantage. Further, the trends are almost similar on the `Best' evaluation set as can be seen from Figure \ref{fig:figure2}.

\section{Conclusion}
 We approach the problem of ranking clarification questions from a natural language inference perspective, which is in contrast to the previous approaches in existing literature (decision-theoretic, generative, etc.) and demonstrate that BERT representations pre-trained on SNLI and MultiNLI can achieve very high performance on the task. We also showed that the context provided by the answer isn't helpful in learning representational similarities  for the task.

\bibliographystyle{ACM-Reference-Format}

\bibliography{sample-base}

\end{document}